\documentclass[10pt,twocolumn,letterpaper]{article}

\usepackage{cvpr}
\usepackage{times}
\usepackage{epsfig}
\usepackage{graphicx}
\usepackage{amsmath}
\usepackage{amssymb}
\usepackage[bottom]{footmisc}
\usepackage[T1]{fontenc} 
\usepackage{amsfonts}
\usepackage{array} 
\usepackage{xcolor}
\usepackage{multirow}
\usepackage{colortbl} 
\usepackage{indentfirst}
\usepackage{subcaption}
\usepackage{multirow}
\usepackage{footnote}

\usepackage[pagebackref=true,breaklinks=true,letterpaper=true,colorlinks,bookmarks=false]{hyperref}

\begin{document}

\title{Optimal-Landmark-Guided Image Blending for Face Morphing Attacks}

\author{Qiaoyun He, Zongyong Deng, Zuyuan He, Qijun Zhao\thanks{Corresponding author\\This work is supported by the National Natural Science Foundation of China (No. 62176170, 61971005).}\\
College of Computer Science\\
Sichuan University, Chengdu, China\\
{\tt\small \{heqiaoyun1, dengzongyong, hezy\}@stu.scu.edu.cn    qjzhao@scu.edu.cn} 
}

\maketitle
\thispagestyle{empty}

\begin{abstract}
  In this paper, we propose a novel approach for conducting face morphing attacks, which utilizes optimal-landmark-guided image blending. Current face morphing attacks can be categorized into landmark-based and generation-based approaches. Landmark-based methods use geometric transformations to warp facial regions according to averaged landmarks but often produce morphed images with poor visual quality. Generation-based methods, which employ generation models to blend multiple face images, can achieve better visual quality but are often unsuccessful in generating morphed images that can effectively evade state-of-the-art face recognition systems~(FRSs). Our proposed method overcomes the limitations of previous approaches by optimizing the morphing landmarks and using Graph Convolutional Networks (GCNs) to combine landmark and appearance features. We model facial landmarks as nodes in a bipartite graph that is fully connected and utilize GCNs to simulate their spatial and structural relationships. The aim is to capture variations in facial shape and enable accurate manipulation of facial appearance features during the warping process, resulting in morphed facial images that are highly realistic and visually faithful. Experiments on two public datasets prove that our method inherits the advantages of previous landmark-based and generation-based methods and generates morphed images with higher quality, posing a more significant threat to state-of-the-art FRSs.

\end{abstract}   

\section{Introduction}

Facial features are widely used in various security applications, especially for border control, due to the advantages of non-contact and fast recognition speed. Although the security of face recognition systems~(FRSs) is constantly improving, the recently emerging face morphing attack poses a significant threat to scenarios like Automatic Border Control~(ABC). It aims to generate a single face image by combining two or more images. The resulting image, which we refer to as the morphed image, contains the biometric features of the contributing subjects and can be successfully verified to multiply identities~\cite{ferraraMagicPassport2014}. Face morphing attacks would lead to security vulnerability of FRSs and widespread criminal activities.

\begin{figure*}[t]
  \centering
  \includegraphics[scale=0.41]{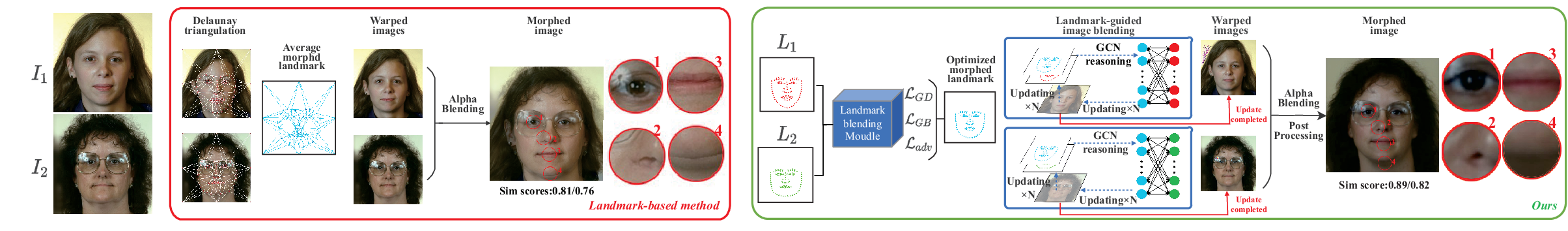}

  \caption{Compared to existing landmark-based methods, our proposed approach can generate high-quality morphs while also increasing the vulnerability of facial recognition systems. Traditional landmark-based methods, as depicted in the red box, partition the face into multiple triangles, resulting in discontinuities and unnatural phenomena at the connections between triangles in fine-detail areas (red circles 2, 3). Additionally, inaccurate landmark detection or missing landmarks can lead to duplicated images in the eye region (red circle 1), and jagged edges can appear along the edges of the face (red circle 4). In contrast, our landmark blending module enhances the similarity between the morphed and original images by optimizing the morphed landmarks. Furthermore, we use GCNs to analyze the relationships between landmarks, minimizing facial artifacts in the generated morphed images. (Arcface~\cite{DBLP:conf/cvpr/DengGXZ19} FRS computes the similarity score to measure how similar a morphed image is to the two original images it was created from.)}
  \label{fig:method}
  \end{figure*} 

The key to effectively researching and defending against morphing attacks is to generate high-quality morphed images, which should match all the contributing subjects successfully. Existing morphing generation methods can be divided into two categories, namely generation-based methods and landmark-based methods. Generation-based morphing methods generate morphs by learning the latent appearance representations of contributing images and blending them in the latent space~\cite{DBLP:conf/iwbf/VenkateshZRRDB20}. However, a major drawback of this approach is the inability to preserve the identity of the contributing images. While these methods can generate high-quality images by blending the latent appearance representations, they lose the critical details and features of the original images, resulting in morphs that fail to pass FRSs' verification. Landmark-based morphing methods involve the detection of facial landmarks in contributing images. Subsequently, contributing images are warped to align the positions of average landmarks, followed by alpha blending to produce the morphed image~\cite{ferraraDecouplingTextureBlending2019}. Typically, Delaunay triangulation is used to divide the facial space between landmarks into triangles for efficient image warping. However, this method has several drawbacks that need to be considered: \textbf{(1) Discontinuous and unnatural appearance}: Due to the Delaunay triangulation dividing the image into triangles, the edges between triangles may result in a discontinuous and unnatural appearance particularly in some detailed areas such as the nose or mouth, as shown in Fig.~\ref{fig:method}. Moreover, in some regions of the face, such as the chin, the triangles may not align perfectly with the shape of the face, creating sharp edges that can appear jagged or unnatural in the final morphed image. \textbf{(2) Dependence on precise landmark detection}: The accuracy and realism of landmark-based morphing rely heavily on precise landmark detection. Significant artifacts can appear in the eye area if the landmarks are not detected accurately, which can affect the quality of the image. \textbf{(3) Distortions and loss of identity information}: Averaging landmarks to create an intermediate shape in landmark-based morphing may cause distortions and loss of identity information, especially if the faces have vastly different structures. This can lead to poor recognition and unsatisfactory results in the final morphed image.
 
\raggedbottom
To address the limitations of existing morphing methods, we propose an optimal-landmark-guided image blending to achieve a better balance between visual quality and the attack success rate of morphed images. Specifically, instead of simply averaging contributing landmarks, we propose a landmark blending module to compute optimal morphed landmarks with a more sophisticated and fine-grained algorithm, thus better preserving the facial geometry of the contributing subjects. With the obtained morphed landmarks, we further propose a landmark guided image blending module for generating natural-looking morphed images. The proposed module specifically utilizes Graph Convolutional Networks (GCNs) to fully exploit the spatial and structural differences between the contributing subjects and optimal landmarks, guiding the update of image appearance features. To achieve this, we treat each point in the two landmarks as a separate node and connect them to a fully connected bipartite graph. This graph structure enables the extraction of geometric features and relationships between the nodes using GCNs. By leveraging the ability of GCNs to capture the dependencies between landmarks and guide the precise modification of appearance features around each landmark during the warping process, our approach effectively avoids visual artifacts that commonly arise from traditional Delaunay triangulation-based warping methods that rely on triangle subdivision for guidance. Compared to existing methods, our approach enables smooth face shape warping from one identity to another, with fewer visual artifacts and better preservation of identity. We conduct qualitative and quantitative experiments on the FERET~\cite{FERETa} and FRGC-V2~\cite{DBLP:conf/cvpr/PhillipsFSBCHMMW05} datasets, and the results show that the proposed method poses a greater threat to FRSs while improving the visual quality of the morphed images to deceive human experts.

The remainder of this paper is organized as follows. Section~\ref{2} briefly reviews related work. Section~\ref{3} introduces our proposed method in detail. Section~\ref{4} then reports the experimental results. Section~\ref{5} finally concludes the paper. 

\begin{figure*}[t]\footnotesize
  \centering
  \includegraphics[scale=0.430]{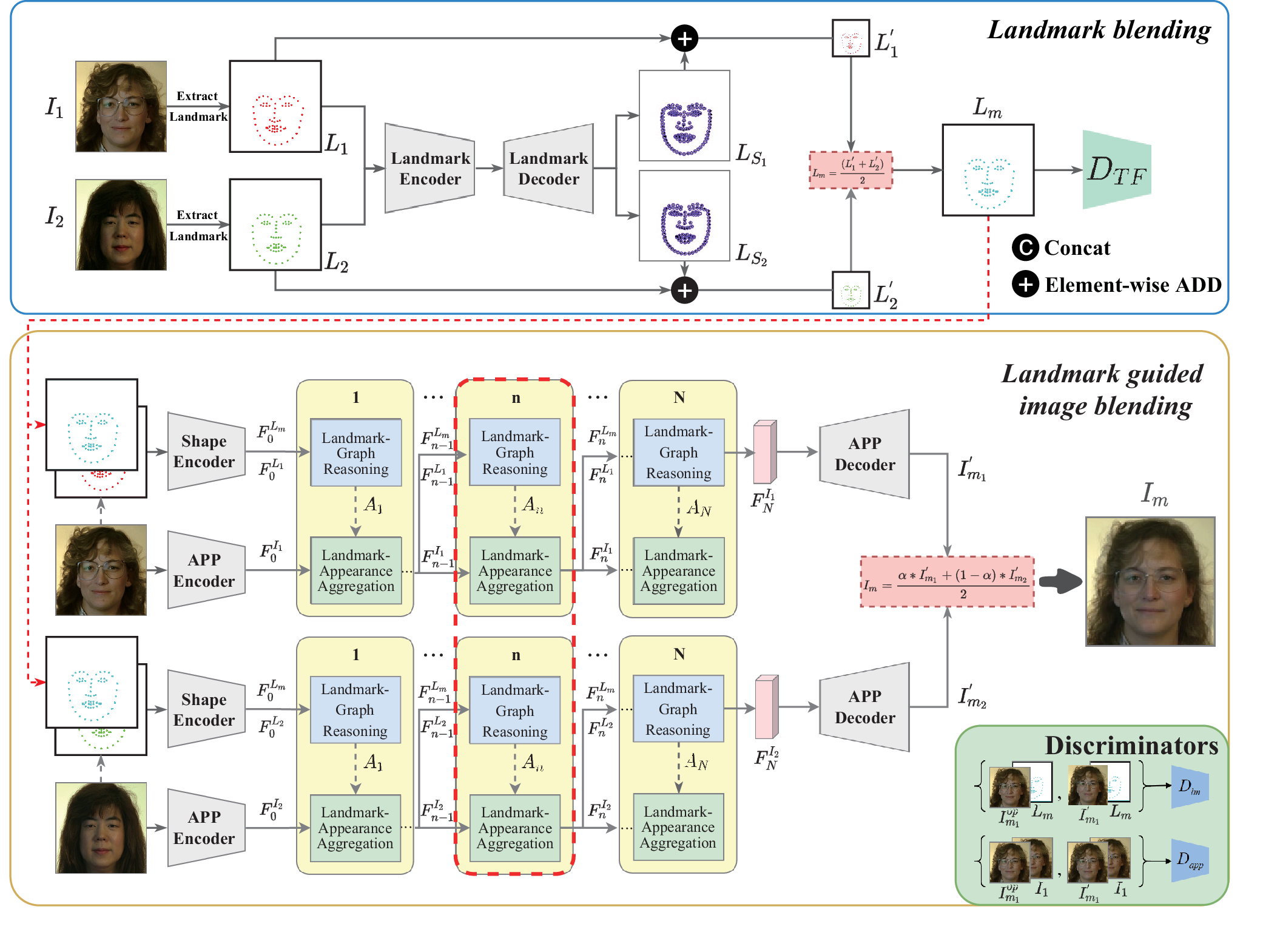}

  \caption{Workflow of Optimal Landmark-Guided Morph Generation. Our proposed method consists of two modules: landmark blending and landmark guided image blending. The former is designed to identify optimized morphed landmark $ L_m $ that preserves facial structures similar to both contributing images. The morphed landmarks guide the blending process of facial appearances in the landmark guided image blending module. To accomplish this, we employ GCNs to naturally warp the original image by inferring the relationships between the morphed landmarks $L_m$ and the contributing original landmarks $L_1$ and $L_2$. Following $N$ iterations of GCN inference, the intermediate warping images $I^{'}_{m_1}$ and $I^{'}_{m_2}$ are combined to produce the final morphed image.}
  \label{fig:method2}
  \end{figure*}

\section{Related Work}\label{2}

\subsection{Landmark-based morphing}


In a landmark-based morphing generation, the facial landmarks of the two contributing images are first detected, \textit{e.g.}, by using Dlib library, and the average position of the landmarks is then calculated. Afterwards, the source image is warped to align with the target landmarks. One approach for image warping is to use the average landmarks to create a Delaunay triangulation. This triangulation is then employed to warp the source image by applying affine transformations to each triangle. The resulting intermediate image closely approximates the target image. Finally, these warped images are blended according to the blending factor to generate the morphed images. Typically, the average landmarks used for morphing are obtained by linear interpolation of the landmarks of the contributing images. However, if the difference in facial features between the contributing images is excessively large or some landmarks are misaligned, this approach may result in unreliable facial geometry. Several studies have exploited the geometric inconsistency between morphed landmarks and real landmarks for morphing attack detection~\cite{autherithDetectingMorphingAttacks2020,damerDetectingFaceMorphing2019}. In such cases, more sophisticated techniques, such as nonlinear optimization or machine learning algorithms, may be necessary to obtain accurate morphing results.


 \subsection{Guided image-to-image generation}

The use of landmarks, human skeletons and scene segmentation maps to guide image generation has been proposed in various computer vision tasks~\cite{tangTotalGenerateCycle2022}. Among them, the task of human pose-guided image generation is highly relevant to our work. The key to this task is to infer the remote relationship between the source pose and the target pose. Most studies relied on convolutional layers~\cite{DBLP:journals/corr/abs-2211-06719} to model local relationships between source and target poses. Instead, BiGraphGAN~\cite{DBLP:journals/corr/abs-2211-06719} projects the source and target poses into a bipartite graph and infers the cross-long-range relationship between the poses in the bipartite graph via GCNs. Inspired by this work, we propose to use the GCNs for face image morphing. Specifically, we exploit the relationship between facial landmarks and capture changes in facial details using the graph structure. Our approach enables us to achieve more natural morphs compared to traditional methods. 

\raggedbottom
\section{Methodology}\label{3}

In this section, we presented our proposed method for improving the quality of morphed images while ensuring the ability to successfully bypass FRSs. Our method consists of two key modules: a landmark blending module and a landmark guided image blending module. The landmark blending module employs one encoder and one decoder to calculate optimal morphed landmarks. On the other hand, the landmark blending module employs GCNs to capture the spatial and structural information of two distinctive landmarks, thereby facilitating the learning of the interrelationships among facial landmarks. By doing so, this module enables the identification of facial regions that require modification to achieve smoother facial shape warping while improving the preservation of identity compared to existing landmark-based methods. Finally, we discussed the implementation details of our proposed method.

\subsection{Landmark blending}

In contrast to traditional methods that use linear interpolation between landmarks, our proposed method employs a landmark blending module to generate more precise and sophisticated morphed landmarks. Fig.~\ref{fig:method2} illustrates this module, which includes one landmark encoder and one landmark decoder. The encoder is used to extract information about the geometric structure of the input landmarks $L_1$ and $L_2$, and the decoder uses this information to estimate the landmark shifts required to produce the final morphed landmark. The landmark shifts $L_{S_1}$ and $L_{S_2}$ represent the displacement of $L_1$ with respect to $L_2$, and $L_2$ with respect to $L_1$, respectively. These shifts are applied in a point-wise manner to the original landmarks $L_1$ and $L_2$, resulting in the transformed landmarks $L_1^{'}$ and $L_2^{'}$. These shifted landmarks preserve the facial contour of one input and the pose information of the other. Finally, the blended landmarks $L_m$ are obtained as the average of $L_1^{'}$ and $L_2^{'}$.
\begin{equation}
	L_m=\frac{(L_1^{'}+L_2^{'})}{2}
      =\frac{(L_1+L_{S_1})+(L_2+L_{S_2})}{2}.
\end{equation}

The proposed method employs $L_{S_i}~(i=1, 2)$ to capture the mutual influence between the two input landmarks, thereby achieving more accurate facial shape transformation and optimized landmarks. The transformed landmarks $L_1^{'}$ and $L_2^{'}$ have distinctive attributes of both $L_1$ and $L_2$ that can be synthesized by averaging them to obtain $L_m$, further improving the naturalness and accuracy of the morphed landmarks.

During the training phase of the landmark blending module, we employ a set of loss functions to guide the optimization process. These loss functions include the geometric distance loss ($\mathcal{L}_{GD}$), the geometric balance loss ($\mathcal{L}_{GB}$), and the adversarial loss ($\mathcal{L}_{adv}$). 

\noindent\textbf{Geometric distance loss}. The first term $\mathcal{L}_{GD}$ is defined by the $ l_1 $ distances between the morphed landmarks and the original contributing landmarks, which can be expressed as:
\begin{equation}
	\mathcal{L}_{GD}=\frac{\left \|L_m-L_1 \right \|_1+\left \|L_m-L_2 \right \|_1 }{2}.
\end{equation}
The objective of $\mathcal{L}_{GD}$ is to minimize the dissimilarity between the morphed and contributing landmarks.

\noindent\textbf{Geometric balance loss}. The second term $ \mathcal{L}_{GB} $  on the principle that the structural similarity between the morphed image and the contributing images should be balanced to avoid bias towards either image. By reducing this imbalance, the resulting morphed image is more likely to exhibit high facial structural similarity to both contributing images, which is crucial for ensuring successful verification in FRSs. The $ \mathcal{L}_{GB} $ can be expressed as:
\begin{equation}
	\mathcal{L}_{GB}=\left |\left \|L_m-L_1 \right \|_1-\left \|L_m-L_2 \right \|_1    \right |.
\end{equation}

\noindent\textbf{Adversarial loss}. The third term of the loss function introduces a landmark discriminator $D_{TF}$ to enforce consistency between the generated facial landmarks and the distribution of real landmarks. Essentially, the landmark blending module acts as a generator $ G $, while the discriminator evaluates the realism of the generated landmarks. The adversarial loss is as follows:
\begin{equation}
  \begin{aligned}
    \mathcal{L}_{adv}=\min_{G} \max_{D_{TF}}~&\mathbb{E}_{L \sim p_{data}(L)}\left[\log \left(D_{TF}(L)\right)\right]+\\
    & \mathbb{E}_{L_z \sim p_{data}(L_z)}\left[\log \left(1-D_{TF}(G(L_z))\right)\right],
    \end{aligned}
\end{equation}
where $L$ is the real landmarks data and $L_z$ denotes the input landmarks data of $G$.

The total loss function of the landmark blending module is given by:
\begin{equation}
  \mathcal{L}_{total}=\lambda_1\mathcal{L}_{GD}+\lambda_2\mathcal{L}_{GB}+\lambda_3\mathcal{L}_{adv},
   \end{equation}
where $\lambda_i$ $(i=1,2,3)$ denotes the hyperparameter of each of the three components. To train the module from scratch, we set the corresponding loss weights as $\lambda_1=100$, $\lambda_2=10$, and $\lambda_3=0.1$.

\subsection{Landmark guided image blending}

\subsubsection{Morphed image generation}

Once the morphed landmarks $L_{m}$ are obtained, we can then use them to guide the generation of morphed image $I_{m}$ from the contributing face images $I_{1}$ and $I_{2}$. For the face image $ I_1 $, it is fed to the appearance encoder to obtain its initial appearance code $ F^{I_1}_0 $, while the landmarks $ L_1 $ and $ L_m $ are fed to the shape encoder to obtain their initial shape codes $F^{L_1}_0 $ and $F^{L_m}_0 $. The initial appearance code and shape code of face image $ I_2 $  are obtained in a similar way to $ I_1 $. 

As shown in Fig.~\ref{fig:method2}, the Landmark guided image blending module comprises of a sequence of Landmark-Graph Reasoning blocks and Landmark-Appearance Aggregation blocks. The Landmark-Graph Reasoning blocks aim to deduce the interdependencies among shape codes to facilitate the modification of appearance codes. Conversely, the Landmark-Appearance Aggregation blocks are employed to sustain the alignment between shape codes and appearance codes throughout the updating process. The individual procedures of these blocks are delineated below.

\textbf{Landmark-Graph Reasoning}. To illustrate this process, take contributing subject $I_1$  as an example. Considering the $n$-th  block shown in Fig.~\ref{fig:method2}, the shape codes $ F^{L_m}_{n-1} $ and $  F^{L_1}_{n-1} $  are concatenated and fed into the $n$-th landmark-graph reasoning block, which uses GCNs to extract differences between the contributing and morphed landmarks. To achieve this, a fully connected bipartite graph is constructed by representing each point in the contributing and morphed landmarks as a node and connecting them to form two distinct sets. The relationships between corresponding points are captured by the interconnectivity between nodes of different sets. By deploying GCNs on this graph structure, both local and global relationships between landmarks are captured, leading to precise and comprehensive modeling of shape differences that exist between the contributing and morphed landmarks. The process can be formulated as follows:

\begin{equation}
A_{n}=\sigma (GCN(Concat(F^{L_m}_{n-1}, F^{L_1}_{n-1} ))),
\end{equation}
where $\sigma(\cdot)$ denotes the Sigmoid function to generate an attention map, highlighting the subtle changes between the contributing and morphed landmarks, as represented by the differences in the graph structure. 

\textbf{Landmark-Appearance Aggregation}. The Landmark-Appearance Aggregation block then uses the obtained attention map $ A_{n} $ to guide the updating of appearance code by:
\begin{equation}
  F^{I_1}_n=A_{n}\otimes F^{I_1}_{n-1}+ F^{I_1}_{n-1},
  \end{equation}
where $ \otimes $ denotes element-wise production. The appearance code $F^{I_1}_{n-1}$ is multiplied with the attention map $A_{n}$ to preserve or suppress features of certain areas to obtain the updated appearance. This process allows the updated appearance code to capture the differences between the contributing and morphed objects in a more precise and localized manner.

In order to ensure that the shape and appearance codes remain synchronized, when the appearance code $F^{I_1}_{n}$ is updated using Eq.~(7), the shape codes $F^{L_m}_{n-1}$ and $F^{L_1}_{n-1}$ are also updated. Specifically, the shape codes are concatenated with the updated appearance code and then fed to convolutional layers. The outputs of these layers are the updated shape codes $F^{L_m}_{n}$ and $F^{L_1}_{n}$, which determine how the facial landmarks should be transformed based on the new appearance information. By synchronizing the updates of shape and appearance codes throughout the network, the model can consistently learn and integrate the information from the morphed landmarks, resulting in the generation of high-quality facial images that are both realistic and consistent. This procedure can be formulated as follows:
\begin{equation}
   F^{L_m}_{n}, F^{L_1}_{n}=Conv(Concat(F^{I_1}_{n}, F^{L_m}_{n-1}, F^{L_1}_{n-1})).
   \end{equation}

\textbf{Generating the final morphed image}. Such operation repeatedly performs for $ N $ iterations to obtain the final appearance code $F^{I_1}_N $, which is then fed to the appearance decoder to generate the intermediate morphed image $ I^{'}_{m_1} $ of contributing subject one. The use of multiple iterations facilitates the gradual optimization of both appearance and shape codes, leading to an approach toward the optimal solution. The employment of multiple iterations is necessary, as a single update may not be sufficient for achieving the optimal solution, thereby requiring continuous refinement for improved outcomes. Moreover, the concurrent update of both appearance and shape codes in each iteration enables the full consideration of their mutual influence, thus yielding more realistic and consistent facial images.

We follow the same procedure as mentioned above to obtain $ I^{'}_{m_2} $ corresponding to the second contributing subject $ I_2 $. The final morphed image is obtained by averaging the $ I^{'}_{m_1} $ and $ I^{'}_{m_2}$:
\begin{equation}
I_m=\frac{\alpha *I^{'}_{m_1}+(1-\alpha)*I^{'}_{m_2}}{2},
\end{equation}
where $ \alpha $ denotes the morphing factor, which is typically set as $ \alpha=0.5 $.
\subsubsection{Adversarial training}

\indent\textbf{Discriminators}. In this paper, we employ two discriminators, the landmark discriminator and the appearance discriminator, to train the Landmark guided image blending module in a manner of adversarial learning. To enhance the discriminative ability of the two discriminators, we introduce the image pairs as the input to the discriminators. Taking image $ I_1 $ as an example, the landmark discriminator takes as input image-landmark pairs, namely $ (I^{op}_{m_1}, L_m) $ and $ (I^{'}_{m_1}, L_m)$, while the appearance discriminator takes image-image pairs, namely $ (I^{op}_{m_1}, I_1) $ and $ ( I^{'}_{m_1}, I_1,) $. Here, $ I^{op}_{m_1} $ refers to the morphed image of $ I_1 $ obtained based on $ L_m $ using the OpenCV-based method. While we still use OpenCV-generated morphed images as a supervisor, our method avoids the undesirable artifacts that can result from using Delaunay triangulation. By leveraging the powerful image generation capabilities of GANs, we are able to produce high-quality morphed images with a nearly identical facial structure to the OpenCV-generated images but with fewer artifacts. Training a GAN using paired image samples strengthens the discriminator's ability to distinguish between real and fake images, which improves the overall quality of the generated images. Using paired image samples also speeds up the convergence of the training process, resulting in quicker generation of high-quality images.

\textbf{The loss function}. 
We aim to optimize our model using multiple objectives, including perceptual loss, pixel-wise $ l_1 $  loss, Multi-Scale Structural Similarity (MS-SSIM) loss, and adversarial loss. The proposed loss function can be formulated as follows:

\begin{equation}
  \mathcal{L}_{total}=\lambda_1\mathcal{L}_{per}+\lambda_2\mathcal{L}_{l_1}
   +\lambda_3\mathcal{L}_{MS-SSIM}+\lambda_4\mathcal{L}_{adv}.
\end{equation}

The hyperparameters $\lambda_i$ $(i=1, 2, 3, 4)$ represent the individual parameters associated with the four components, where $\lambda_1 = 10$, $\lambda_2 = 10$, $\lambda_3 = 1$, and $\lambda_4 = 5$.

\subsection{Implementation details}
In all our experiments, we use the morphed image pairs described in Section~\ref{4.1} for training. For each face image, we first extract 106 landmarks and then crop and align the face image according to the FFHQ~\cite{{DBLP:journals/corr/abs-1812-04948}} process. For the landmark blending module, we use the Adam optimizer to optimize the encoder and decoder, and the initial learning rate is set to $ 3e^{-4} $. For the Landmark guided image blending module, we also use the Adam optimizer, and the initial learning rates of the generator and the discriminator are both set to $ 2e^{-4} $, while the number of iterations $ N $ is set to $ N=9 $.

\section{Experimental results}\label{4}

\subsection{Datasets and evaluation criteria}\label{4.1}

\textbf{Datasets setup}. FERET~\cite{FERETa} and FRGC-V2~\cite{DBLP:conf/cvpr/PhillipsFSBCHMMW05} are standard datasets commonly used in the literature of morph generation and detection. They contain a large number of images with different identities. To effectively evaluate the quality of generated morphed images with comparison to earlier work, we introduced the FERET-Morphs and FRGC-Morphs datasets~\cite{sarkarVulnerabilityAnalysisFace2020}, which contain four types of morphing attacks generated by OpenCV\cite{Facea}, FaceMorpher~\cite{FaceMorpher}, StyleGAN2~\cite{DBLP:conf/iwbf/VenkateshZRRDB20}, MIPGAN-II~\cite{DBLP:journals/tbbis/ZhangVRRDB21}. The FERET-Morphs dataset contains 529 morphs and the FRGC-Morphs dataset contains 964 morphs for each morph type. We followed the protocol in \cite{DBLP:journals/tifs/ScherhagRMB20} and selected the same morphing pairs as in \cite{sarkarVulnerabilityAnalysisFace2020}.

\textbf{Evaluation criteria} \emph{MMPMR}. In our experiments, we use the metric Mated Morph Presentation Match Rate~(MMPMR) to evaluate the vulnerability of FRSs to morphing attacks. MMPMR is calculated by comparing the similarity scores of each morphed image with the images of the two contributing subjects. The minimum similarity score is compared with a fixed threshold to discriminate the success of the morphing attack.
\begin{equation}
  \operatorname{MMPMR}(\tau)=\frac{1}{M} \cdot \sum_{m=1}^M\left\{\left[\min _{n=1, \ldots, N_m} S_m^n\right]>\tau\right\}.
\end{equation}    
Following the guidelines of Frontex~\cite{Best}, here $\tau$ is the corresponding verification threshold of FRS at FAR = $0.001 (\%)$. $M$ is the number of morphed images, $ N_m $  is the number of subjects contributing to morph $ m $, and $ S^n_m $ is the matched similarity score between morph $ m $  and the $ n $-th contributing subject. Obviously, Higher MMPMR values heighten the threat of morphing attacks against FRSs, \textit{i.e.}, the quality of generated morphed images is positively associated with the level of risk.

\begin{table}[t]\footnotesize
  \centering
  \caption{MMPMR@FMR = $0.1 \%$ of different morphing attack methods against five FRSs on FERET and FRGC-V2.}
    \begin{tabular}{|c|c|c|c|}
    \hline
    \multicolumn{1}{|c|}{FRS} & Methods & FERET & FRGC-V2 \\ \hline
    \hline
    \multirow{5}{*}{Arcface~\cite{DBLP:conf/cvpr/DengGXZ19}} & OpenCV\cite{Facea}& 34.6& 77.0 \\
          & FaceMorpher~\cite{FaceMorpher}& 34.1& 74.6 \\
          & StyleGAN2~\cite{DBLP:conf/iwbf/VenkateshZRRDB20}& 2.4& 23.9 \\
          & MIPGAN-II~\cite{DBLP:journals/tbbis/ZhangVRRDB21}& 26.0& 63.5 \\
          & Ours & \textbf{48.4}& \textbf{78.3} \\
    \hline
    \multirow{5}{*}{FaceNet~\cite{schroffFaceNetUnifiedEmbedding2015}} & OpenCV\cite{Facea}& 41.1& 71.1 \\
          & FaceMorpher~\cite{FaceMorpher}& 39.9& 70.7 \\
          & StyleGAN2~\cite{DBLP:conf/iwbf/VenkateshZRRDB20}& 1.6& 9.6 \\
          & MIPGAN-II~\cite{DBLP:journals/tbbis/ZhangVRRDB21} & 32.9& 63.0 \\
          & Ours & \textbf{43.1}& \textbf{72.4}\\
    \hline
    \multirow{5}{*}{VGG~\cite{Deep}} & OpenCV\cite{Facea}& 22.0& 61.9\\
          & FaceMorpher~\cite{FaceMorpher}& 20.5& 61.3 \\
          & StyleGAN2~\cite{DBLP:conf/iwbf/VenkateshZRRDB20}& 2.0 & 10.1  \\
          & MIPGAN-II~\cite{DBLP:journals/tbbis/ZhangVRRDB21}& 14.5& 56.7 \\
          & Ours & \textbf{24.8}& \textbf{62.4} \\
    \hline
    \multirow{5}{*}{ISV-based~\cite{DBLP:conf/icb/WallaceMMM11}} & OpenCV\cite{Facea}& 44.8&96.9 \\
          & FaceMorpher~\cite{FaceMorpher}& 42.6& 95.6 \\
          & StyleGAN2~\cite{DBLP:conf/iwbf/VenkateshZRRDB20}& 2.7 & 17.5\\
          & MIPGAN-II~\cite{DBLP:journals/tbbis/ZhangVRRDB21}& 7.3 & 30.0 \\
          & Ours & \textbf{45.6}&\textbf{97.4} \\
    \hline
    \end{tabular}%
  \label{tab:table1}%
\end{table}%

\emph{APCER and BPCER}. We evaluate the detectability of generated morphed images by assessing the robustness of the morphing attack detection (MAD) technique using the following metrics: Attack Presentation Classification Match Rate (APCER) and Bona fide Presentation Classification Match Rate (BPCER). The APCER defines the proportion of attack samples that are misclassified as bona fide face samples, and conversely, BPCER defines the proportion of bona fide face samples that are misclassified as attack samples.

\emph{PSNR and SSIM}. We present the quantitative evaluation results of our proposed morphed image generation techniques using two perceptual image quality metrics: Peak Signal-to-Noise Ratio (PSNR) and Structural Similarity Index (SSIM). These metrics are calculated based on the reference image, which is constructed from parent face images of two data subjects. Specifically, we used the parent face images from both data subjects as the reference image to assess the generated morphed image. The image quality scores were averaged across both parent images.

\subsection{Results and analysis}
\subsubsection{Comparision with State-of-the-Arts}

\textbf{Attacking FRSs with morphed images}. We employed four FRSs to assess the impact of morphing attacks using morphed images. Among these, three FRSs were based on deep learning models and were sourced from open repositories, while the remaining system served as a ``classical'' baseline. Specifically, the open-source FRSs employed in our study were FaceNet~\cite{schroffFaceNetUnifiedEmbedding2015}, Arcface~\cite{DBLP:conf/cvpr/DengGXZ19}, and VGG~\cite{Deep}. The classical baseline was ISV-based~\cite{DBLP:conf/icb/WallaceMMM11}. All FRSs used the same pre-trained models, and decision thresholds as previously described in \cite{sarkarVulnerabilityAnalysisFace2020}.

Table~\ref{tab:table1} presents a comparison of the MMPMR values of four FRSs using different morphing techniques. Our proposed method surpasses landmark-based methods and achieves the highest threat scores on all FRSs. In contrast to previous studies, which typically showed that generation-based morphing techniques were inferior to landmark-based techniques in terms of MMPMR results, our approach combines both methods to achieve superior visual quality and higher success rates in attacking FRSs.

\begin{table}[ht]\footnotesize
  \centering
  \caption{Detectability of morphed images by MAD (* indicates the data used to train the MAD methods).}
    \begin{tabular}{|c|cc|cc|}
    \hline
    \multirow{3}{*}{Morphs~(Test)} & \multicolumn{2}{c|}{HOG+SVM} & \multicolumn{2}{c|}{MixfaceNet} \\
\cline{2-5}          & \multicolumn{2}{c|}{BPCER(\%)@APCER=} & \multicolumn{2}{c|}{BPCER(\%)@APCER=} \\
\cline{2-5}          & 5\%   & 10\%  & 5\%   & 10\% \\
    \hline
    \rowcolor[rgb]{ .851,  .851,  .851} OpenCV* & 0.00  & 0.00  & 0.94  & 0.00  \\
    FaceMorpher& 0.00  & 0.00  & 0.94  & 0.00  \\
    StyleGAN2~ & \textbf{100.00 } & \textbf{100.00 } & 89.62  & 61.32  \\
    MIPGAN-II~& \textbf{100.00 } & \textbf{100.00 } & 88.68  & 72.64  \\
    Ours  & \textbf{100.00 } & \textbf{100.00 } & \textbf{91.51 } & \textbf{72.64 } \\
    \hline
    OpenCV& 0.00  & 0.00  & 0.00  & 0.00  \\
    \rowcolor[rgb]{ .851,  .851,  .851} FaceMorpher*& 0.00  & 0.00  & 0.00  & 0.00  \\
    StyleGAN2 & \textbf{100.00 } & \textbf{100.00 } & 70.75  & 55.66  \\
    MIPGAN-II& \textbf{100.00 } & \textbf{100.00 } & 83.02  & 65.09  \\
    Ours  & 97.17  & 97.17  & \textbf{96.23 } & \textbf{80.19 } \\
    \hline
    OpenCV& \textbf{100.00 } & \textbf{100.00 } & 90.57  & 84.91  \\
    FaceMorpher& \textbf{100.00 } & \textbf{100.00 } & \textbf{98.11 } & \textbf{94.34 } \\
    \rowcolor[rgb]{ .851,  .851,  .851} StyleGAN2* & 0.00  & 0.00  & 50.94  & 43.40  \\
    MIPGAN-II& 0.00  & 0.00  & 76.42  & 59.43  \\
    Ours  & 50.00  & 38.68  & 97.17  & 83.02  \\
    \hline
    OpenCV& \textbf{100.00 } & \textbf{100.00 } & 93.40  & 85.85  \\
    FaceMorpher& \textbf{100.00 } & \textbf{100.00 } & \textbf{99.06 } & \textbf{98.11 } \\
    StyleGAN2& 0.00  & 0.00  & 55.66  & 48.11  \\
    \rowcolor[rgb]{ .851,  .851,  .851} MIPGAN-II* & 0.00  & 0.00  & 52.83  & 38.68  \\
    Ours  & 18.87  & 9.43  & 78.30  & 63.21  \\
    \hline
    OpenCV& \textbf{98.11 } & \textbf{98.11 } & 90.57  & 88.68  \\
    FaceMorpher& 93.40  & 87.74  & \textbf{92.45}& \textbf{90.57}  \\
    StyleGAN2& 45.28  & 19.81  & 68.87  & 65.09  \\
    MIPGAN-II& 22.64  & 14.15  & 69.81  & 48.11  \\
    \rowcolor[rgb]{ .851,  .851,  .851} Ours* & 0.00  & 0.00  & 81.13  & 53.77  \\
    \hline
    \end{tabular}%
  \label{tab:mad}%
\end{table}%


\begin{table}[htbp]\footnotesize
  \centering
  \caption{Average detection errors for different morphing methods
under unknown attacks. Higher errors mean more severe threats of the morphing methods.}
    \begin{tabular}{|c|c|cc|c|}
    \hline
    \multirow{2}{*}{MAD} & \multirow{2}{*}{Morphs} & \multicolumn{2}{c|}{BPCER(\%)@APCER=} & \multirow{2}{*}{EER(\%)} \\
\cline{3-4}          &       & 5\%   & 10\%  &  \\
    \hline
    \multirow{5}[2]{*}{HOG+SVM} & Opencv & \textbf{74.53} & \textbf{74.53}  & 44.56  \\
          & Facemorpher & 73.35 & 71.94  & \textbf{44.59}  \\
          & StyleGAN2 & 61.32 & 54.95  & 38.68  \\
          & MIPGAN-II & 55.66 & 53.54  & 34.35  \\
          & Ours  & 66.51 & 61.32  & 41.31 \\
    \hline
    \multirow{5}[2]{*}{Mixfacenet} & Opencv & 68.64 & 64.86 & 37.14  \\
          & Facemorpher & 72.64 & 70.76 & 42.73  \\
          & StyleGAN2 & 71.23 & 57.55 & 38.58  \\
          & MIPGAN-II & 79.48 & 61.32 & 37.27  \\
          & Ours  & \textbf{90.8}  & \textbf{74.77} & \textbf{47.59}  \\
    \hline
    \end{tabular}%
  \label{tab:avg}%
\end{table}%

\begin{figure}[t]\footnotesize
  \centering
  \begin{subfigure}[b]{1\linewidth}
    \includegraphics[width=\linewidth]{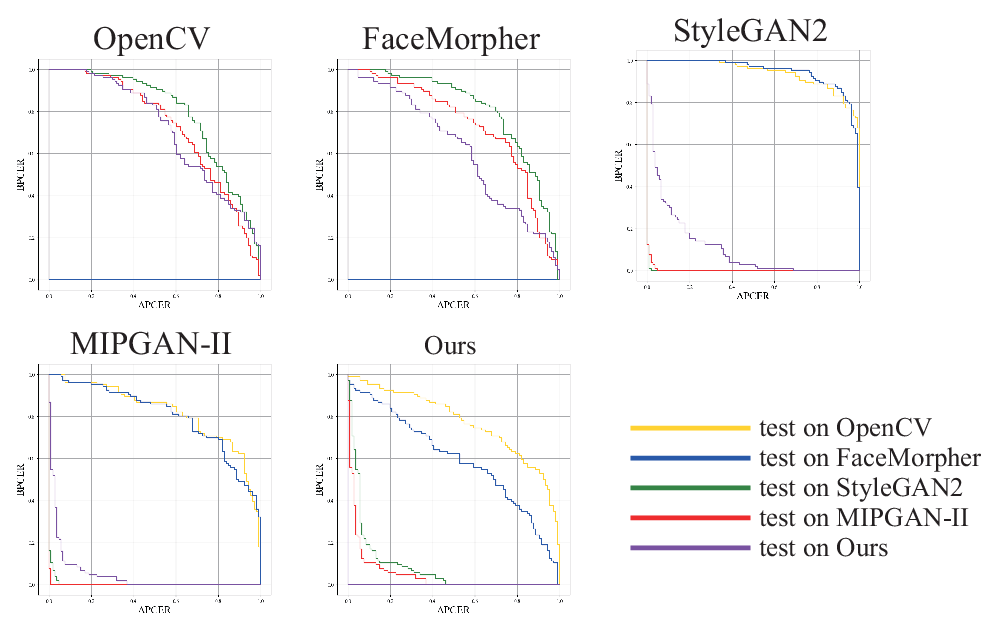}
    \caption{HOG+SVM}
    \label{fig:sub1}
  \end{subfigure}
  \hspace{0.1\linewidth}
  \begin{subfigure}[b]{1\linewidth}
    \includegraphics[width=\linewidth]{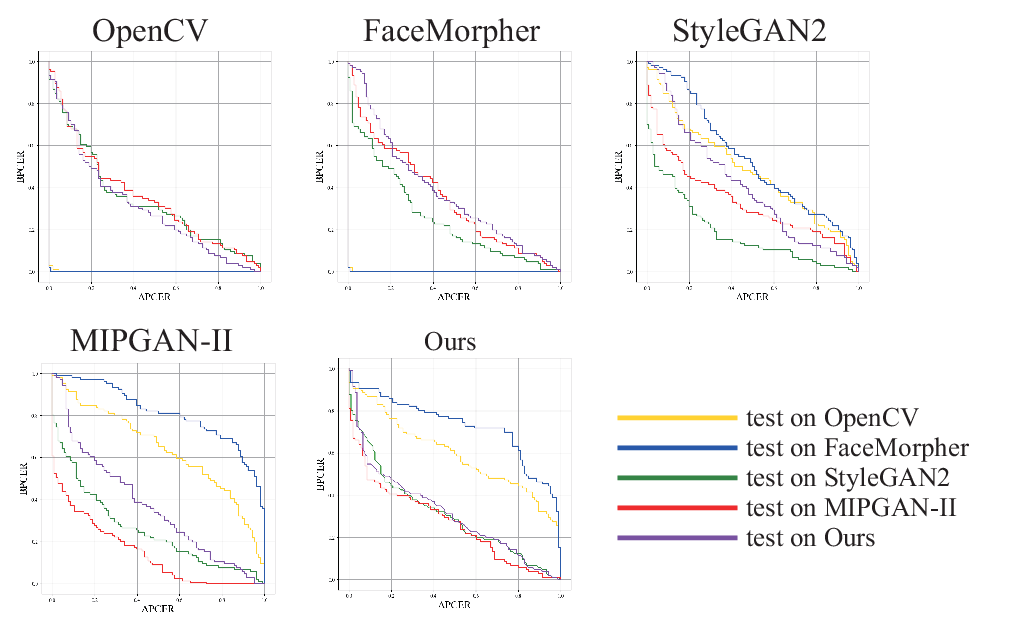}
    \caption{Mixfacenet}
    \label{fig:sub2}
  \end{subfigure}
  \caption{Analysis of DET curves for HOG+SVM and MixfaceNets across diverse test sets and varied training data configurations.}
  \label{fig:det}
\end{figure}

\begin{table}[htbp]\footnotesize
   \centering
   \caption{Quality analysis of morphed images with 95\% confidence interval based on PSNR and SSIM.}
     \begin{tabular}{|c|c|c|}
     \hline
     Methods & PSNR  & SSIM  \\
     \hline
     OpenCV\cite{Facea} & 10.6273 $\pm$ 0.0786 & 0.4092  $\pm$ 0.0063  \\
     \hline
     FaceMorpher~\cite{FaceMorpher} & \hspace{0.5em}9.1290  $\pm$ 0.0972 & 0.3847  $\pm$ 0.0060  \\
     \hline
     StyleGAN2~\cite{DBLP:conf/iwbf/VenkateshZRRDB20}& 14.7230  $\pm$ 0.1490 & 0.6342  $\pm$ 0.0058  \\
     \hline
     MIPGAN-II~\cite{DBLP:journals/tbbis/ZhangVRRDB21}& 14.9792  $\pm$ 0.1592 & 0.6390  $\pm$ 0.0062  \\
     \hline
     Ours  & \textbf{16.6488  $\pm$ 0.1603} & \textbf{0.6729  $\pm$ 0.0056}  \\
     \hline
     \end{tabular}%
   \label{tab:ssim}%
 \end{table}%

 \textbf{Detectability of morphed images by MAD}. To evaluate the detectability of various types of morphing generation methods under known and unknown attack scenarios, we conducted cross-dataset experiments using two state-of-the-art MAD techniques, namely HOG+SVM~\cite{venkateshDetectingMorphedFace2020a} and MixfaceNet~\cite{DBLP:conf/cvpr/DamerLFSPB22}. The BPCER results under different APCER thresholds on the FERET dataset are presented in Table~\ref{tab:mad}. The results highlighted with a gray background in Table~\ref{tab:mad} demonstrate that MAD algorithms exhibit a high level of success in detecting known attacks across various morphing methods, including our proposed method. But unknown attacks~(more common in practical applications) can pass the MAD check with relatively high probability. We have summarized the average detection errors (measured in terms of BPCER and EER) for different morphing methods under unknown attacks in Table~\ref{tab:avg}. Higher MAD errors indicate more severe threats. Notably, our method is one of the techniques that may lead to higher MAD errors. Furthermore, we have presented DET curves in Fig.~\ref{fig:det}, illustrating the performance of HOG+SVM and MixfaceNet on different test sets and training setups. As expected, unknown attacks are harder to detect. In light of these findings, further research and improvements in MAD algorithms are necessary to enhance the detection capabilities and robustness against unknown attacks.

\begin{figure}[t]\footnotesize
  \centering
  \begin{subfigure}[b]{1\linewidth}
    \includegraphics[width=\linewidth]{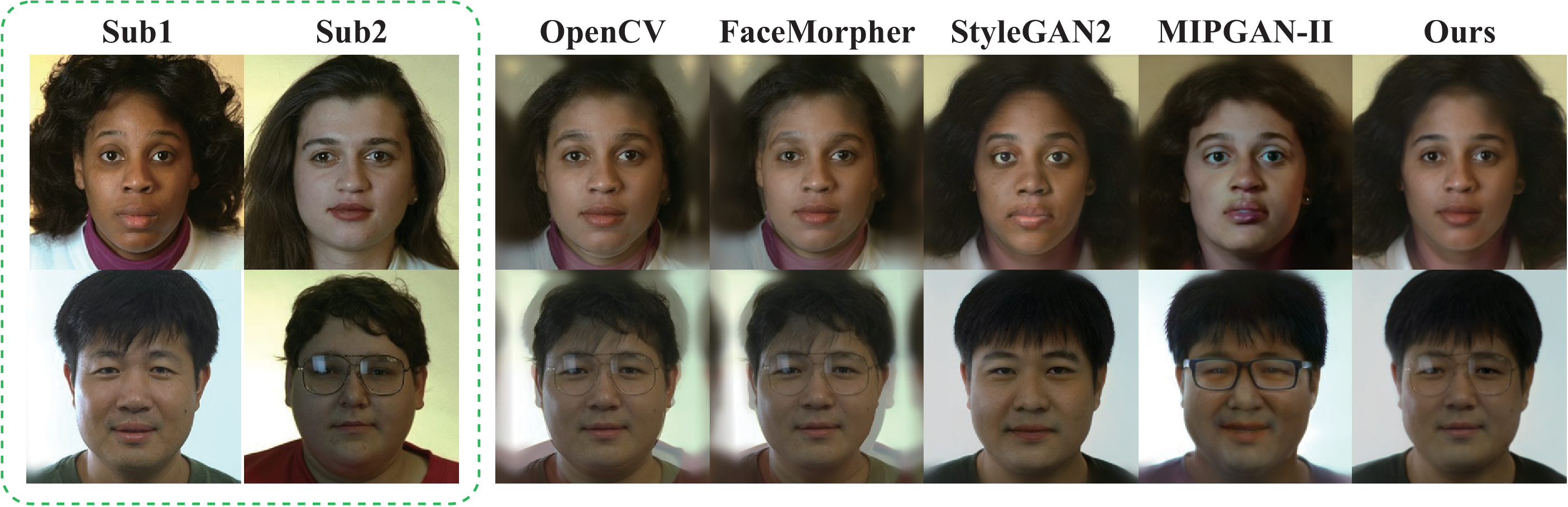}
    \caption{FERET examples}
    \label{fig:sub1}
  \end{subfigure}
  \hspace{0.1\linewidth}
  \begin{subfigure}[b]{1\linewidth}
    \includegraphics[width=\linewidth]{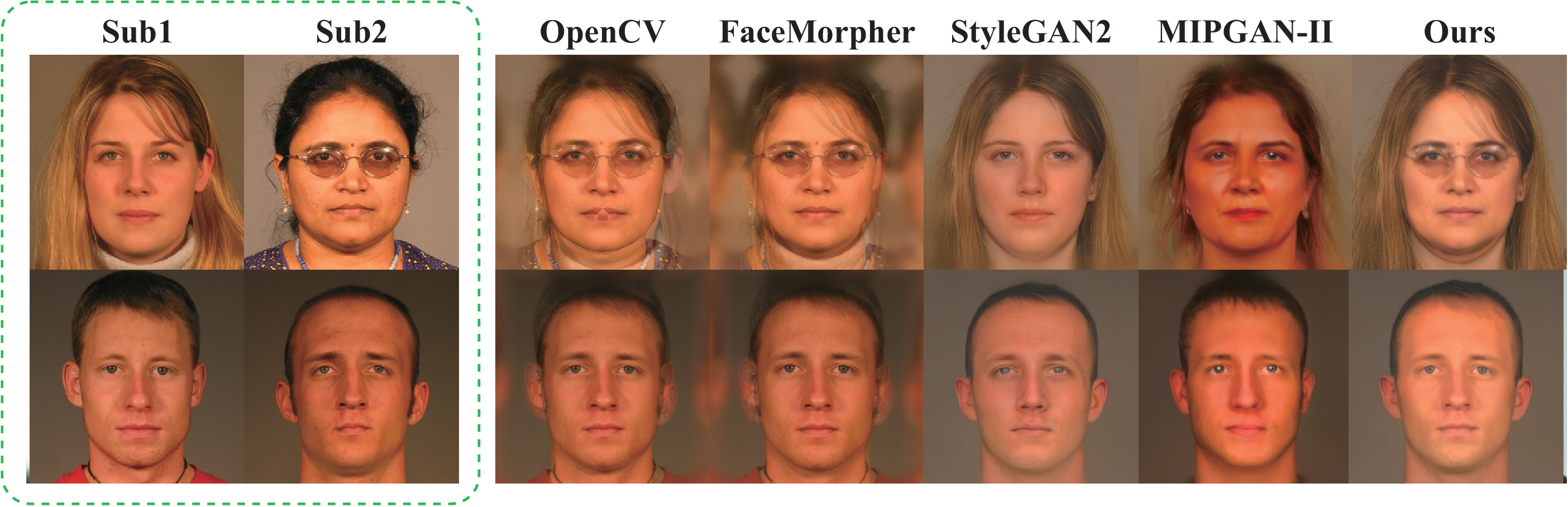}
    \caption{FRGC-V2 examples}
    \label{fig:sub2}
  \end{subfigure}
  \caption{Comparing the Quality of Face Morphs Generated by the Proposed Method and Existing Approaches: A Qualitative Analysis.}
  \label{fig:compare}
\end{figure}

\textbf{Perceptual Image Quality Analysis}. The results presented in Table~\ref{tab:ssim} demonstrate the superior performance of our proposed method compared to existing state-of-the-art morph generation techniques. Our approach achieved higher values of PSNR and SSIM, indicating its effectiveness in enhancing the visual appeal of morphed images and making them appear more natural and visually pleasing. In accordance with the observations illustrated in Fig.\ref{fig:compare}, our proposed method yields morphed images with superior visual quality compared to state-of-the-art competing methods for morphing generation.

\begin{table}[ht]\footnotesize
   \centering
   \caption{Performance evaluation of Landmark Blending Module and discriminators of Landmark guided Image Blending Module using MMPMR@FMR = $0.001 (\%)$ metric on FERET dataset: It should be noted that the abbreviation LBM refers to the Landmark Blending Module.}
     \begin{tabular}{|c|c|c|c|c|}
     \hline
     \multirow{2}{*}{Methods} & \multicolumn{4}{c|}{MMPMR(\%)}  \\
 \cline{2-5}          & Arcface & FaceNet& VGG   & ISV-based\\
     \hline
     Ours (w/o LBM)  & 45.9  & 40.1  & 21.6  & 43.3  \\
     \hline
     Ours (w/o $ D_{app}) $  & 40.6  & 36.9  & 17.4  & 35.5  \\
     \hline
     Ours (w/o $ D_{lm}) $  & 46.5  & 38.6  & 20.0    & 41.6  \\
     \hline
     Ours (w $ D_{trod}) $  & 45.6  & 38.0    & 19.1  & 37.6  \\
     
     \hline
     Ours  & \textbf{48.4} & \textbf{43.1} & \textbf{24.8} & \textbf{45.6}  \\
     \hline
     \end{tabular}%
   \label{tab:ablation}%
 \end{table}%

\subsubsection{Ablation study}

\textbf{Effectiveness of landmark blending module
}. To evaluate the impact of the landmark blending module on the performance of our proposed method, we conducted ablation experiments and measured the results using the MMPMR metric on FERET dataset. The method used in the experiments was the traditional method of averaging landmarks, and the experimental results are presented in the first row of Table~\ref{tab:ablation}.

The experimental results show that FRSs without the landmark blending module in our proposed method exhibit a significant decrease in MMPMR values. This finding further supports the effectiveness of the landmark blending module in generating high-quality facial images while preserving the facial structure of the input contributing images, resulting in a decrease in the efficacy of FRSs.

\textbf{Effectiveness of discriminators in Landmark guided image blending module}. To validate the effectiveness of the discriminators in our Landmark guided image blending module, we conducted three ablation experiments. The first experiment trained the model using only the appearance discriminator, while the second used only the landmark discriminator. The third experiment utilized the traditional discriminator input of images $I^{op}_{m_1}$ and $ I^{'}_{m_1}$. To evaluate the performance of our proposed method, we employed the MMPMR metric on the FERET dataset and compared the results of the three experiments, and the results are presented in Table~\ref{tab:ablation}. The findings revealed that our method achieved superior performance compared to the other three methods, as it achieved the highest MMPMR values. These results further underscore the efficacy and feasibility of our approach, which improves the discriminators' discriminative ability to generate high-quality morphed images. By incorporating both appearance and landmark discriminators, our method effectively avoids the undesirable artifacts that can reduce the attack performance against FRSs and morphing attack detection algorithms while also improving the convergence speed of the training process. 

\section{Conclusion}\label{5}

In this paper, we proposed a new morphing attack method against FRSs. The method adaptively generates morphed landmarks to maintain the facial geometry of contributing subjects better. GCNs are then used to extract morphing features from the morphed and contributing landmarks, which are aggregated with appearance features to generate morphed images of both good visual quality and a high attack success rate. We compare our method quantitatively and qualitatively with state-of-the-art methods. The results demonstrate that our method performs better in terms of identity preservation as well as visual quality.


\bibliographystyle{ieee}
{\bibliography{egpaper_for_review}

\end{document}

